%% file: ms.tex
\documentclass[a4paper, 10pt, conference]{ieeeconf}      

\IEEEoverridecommandlockouts                              
\usepackage{multirow}
\usepackage[colorinlistoftodos,prependcaption,textsize=tiny]{todonotes}

\title{\LARGE \bf
Image-to-image Translation as a Unique Source of Knowledge
}

\author{ \parbox{3 in}{\centering Alejandro D. Mousist\\
         Thales Alenia Space\\
         Tres Cantos, Madrid, Spain\\
         {\tt\small alejandro.mousist@thalesaleniaspace.com}}
         \hspace*{ 0.5 in}
}

\begin{document}

\maketitle
\thispagestyle{empty}
\pagestyle{empty}

\input{chapters/0_abstract}

\input{chapters/1_introduction}

\input{chapters/2_methods}

\input{chapters/3_results}

\input{chapters/4_conclusion}

\addtolength{\textheight}{-12cm}   




\section*{Acknowledgment}
This work was partially supported by the Innovation Cluster of Thales Alenia Space. I would also like to thank Elisa Callejo Luis for her kindly support during this research.


\bibliographystyle{IEEEtran}
\bibliography{ms}

\end{document}

%% file: chapters/0_abstract.tex
\begin{abstract}
Image-to-image (I2I) translation is an established way of translating data from one domain to another but the usability of the translated images in the target domain when working with such dissimilar domains as the SAR/optical satellite imagery ones and how much of the origin domain is translated to the target domain is still not clear enough. 
This article address this by performing translations of labelled datasets from the optical domain to the SAR domain with different I2I algorithms from the state-of-the-art, learning from transferred features in the destination domain and evaluating later how much from the original dataset was transferred.
Added to this, stacking is proposed as a way of combining the knowledge learned from the different I2I translations and evaluated against single models.
\end{abstract}

\begin{keywords}
Image-to-Image translation, synthetic aperture radar (SAR), generative adversarial networks, Sentinel
\end{keywords}

%% file: chapters/1_introduction.tex
\section{Introduction}

The continuous growth of new remote sensing sources, diverse sensor types and the development of recent technologies to analyze and extract the knowledge from this miscellaneous data, have encourage the analysis of the Earth Observation data from different perspectives and the possibility to get around some issues like the lack of labelled samples using complementary data.

As it happens in other fields, the success in the application of machine learning techniques is related to the characteristics of the datasets used to train its models. In supervised learning models, the availability of enough labelled data (in terms of volume, variety, correctness, among other characteristics) is crucial to obtain models with an acceptable performance.

Due to difficulties in seeing and recognizing reference objects in SAR images, the availability of tagged data is often low or non-existent for rare objects. On the other hand, these difficulties, added to the fact that for many classes the presence of instances is low in relation to the size of the analyzed regions, make labelling a dataset from scratch not always an affordable task. 

The reverse situation occurs when it comes to optical images, where the amount of existing training data increases every day and the cost of having to assemble a new dataset is notably lower because the images are easily interpreted by the human eye.

In recent years multiple studies \cite{first, Du2021ExploringTP, Wang2019SARtoOpticalIT, Huang, Sasaki} have focus on the possibility of "translating" SAR data to optical data and vice-versa using image-to-image (I2I) translation, i.e., taking images from one domain and transforming them so they have similar characteristics to images from another domain.

In this scenario, the idea of being able to transfer the knowledge contained in a dataset from one domain to other domains is truly tempting. This can become an effective technique in Few-Shot learning problems as augmentation could be done by transforming source domain datasets into destination domain ones with the learned transformers (i2i translation models) \cite{few-shot-definition}.

However, transferred images from one domain to the other are not real. On one hand, the imaging geometry of SAR sensor is completely different from that of optical sensors. On the other, both the reflected echo of SAR sensors (also called backscatter) and the reflected electromagnetic waves captured by the optical sensors provide different information about physical properties of the Earth's surface.
There is no way of transferring all this information from one domain to the other using pixel to pixel translations thus the i2i model only learns to transfer the images from one domain to images that look similar to the images of the destination domain.

L. Liu and B. Lei \cite{first} concluded that although transferred images cannot be considered as real images due to the lack of information on the origin domain of the destination one, main surface features are also remained in translated images using the original implementation of CycleGAN to translate images between SAR/optical domains.  

I2I has been used for performing image matching in the SAR-optical domains.
In \cite{Du2021ExploringTP}, feature matching loss is applied to the cycle consistency generative adversarial network (CycleGAN) for improving image matching.
In \cite{Huang}, PatchPacGAN is used as discriminator in a modified version of CycleGAN to prevent the generation of samples with little diversity (a phenomenon known as mode collapse). Also, a global feature fusion block is introduced in the transfer module of the generator to extract the global feature by fusing features from all the residual dense blocks. Finally, a structural similarity index measure (SSIM) is added to the loss to reduce the effect of blurred images produced by the L1 loss of CycleGAN.

L. Wang et al. \cite{Wang2019SARtoOpticalIT} proposed a supervised method that combines CycleGAN and pix2pix with a successful experimentation on cloud removal of Sentinel-2 imagery. The supervised version of CycleGAN is proposed to keep both the land cover and structure information. A pixel-level MSE loss that represents the difference between the generated image and the target real image is included to the two mappings of CycleGAN.

In \cite{thermal}, an object detection model is fine-tuned on a stylized version of the COCO dataset for detecting vehicles in thermal cameras. Even though the conceptual idea behind this experiment is similar to ours, the problem is different in many aspects. On one hand, geometry similarities between visible/thermal domains are higher when compared with the optical/SAR ones. On the other, satellite images are different from ground level ones and the problem of detecting objects in both types of images is not the same due to multiple factors such as the size of the objects, resolution of images, perspectives, or the availability of ground truths.

H. Sasaki, C. Willcocks and T. Breckon \cite{Sasaki} used a modified version of CycleGAN for performing a smartly mix-
up data augmentation of the destination domain using samples of the source domain as input. The CycleGAN is modified by using Spectral Normalization in conjunction with gradient penalty to prevent mode collapse and stabilize training and both generators and discriminators are conditioned on the class labels present in the input domain.

While all these works have addressed the i2i translation between different domains and most of them have done so by performing translations between the SAR/optical domains specifically, this work focuses on the analysis of how much of the features present in the latent space is propagated by the different selected methods. This is done indirectly by validating if the translated data from the optical domain is still useful for performing tasks like object detection in the SAR domain.

In addition, an ensemble method for combining the results of the different object detection models is proposed on the destination domain. Although this is not the only way of producing the input for an ensemble algorithm (e.g., test-time augmentation is another option in which the detections of multiple transformations of each image are combined), it is the most natural way of doing it \cite{ensembles_2}.

The main contributions of this work are summarized as follows:
\begin{enumerate}
\item The translation of features from one domain to another is validated using different models from the state-of-the-art of i2i translation. This is achieved empirically by translating an object detection dataset from the origin domain to the destination domain and then evaluating the performance of the resulting object detection models trained with the fake data.
\item An ensemble model created with resulting object detection models is proposed to obtain the best from each i2i translation technique used.
\end{enumerate}

%% file: chapters/2_methods.tex
\section{Methods}

\subsection{Data sources}\label{datasources}
\subsubsection{SAR imagery source}
Sentinel-1 IW Level-1 GRD products are used as source for creating the SAR datasets.
Composite RGB (colour) images are generated using the VH polarization channel for red, VV polarization channel for green and the ratio \textbar VH\textbar/\textbar VV\textbar for blue.
The motivation for using this representation is directly related with the class of the objects tagged as it provides a good contrast in the coastal areas.

\subsubsection{Optical imagery source}
In the case of optical datasets, multi-spectral Sentinel-2 MSI Level-1C products are used.
True colour composite is used for the RGB representation of Sentinel-2 imagery, using visible light bands red (B04), green (B03) and blue (B02) in the corresponding red, green and blue colour channels.

\subsection{Image-to-image translation}
\subsubsection{Unpaired dataset}\label{Unpaired-dataset}
The unpaired dataset for training Image-to-Image translation models is generated by cropping the resulting images of the previous listed sources in small patches. The variety of the images is considered when selecting the source products, trying only to avoid areas covered with snow and cloudy images because in previous experimentation with Image-to-Image translation, resulting models were affected when the presence of clouds or snow was high in the training dataset.
The dataset is composed of 6000 Sentinel-2 and 6000 Sentinel-1 patches of 600x600px on each domain.

The vast Copernicus catalogue provides enough variety and volume for the Sentinel-1 and Sentinel-2 data thus augmentation is not needed for creating the unpaired dataset.

\subsubsection{Model selection and training}\label{i2i}
Different unsupervised Image-to-image translation models are used to perform the translations between the SAR and optical domains. Specifically, implementations of the following models are used for doing such a translation:
\begin{itemize}
  \item CycleGAN\cite{cyclegan}
  \item CUT\cite{cut}
  \item DCLGAN\cite{dclgan}
  \item Council-GAN\cite{council}
  \item MUNIT\cite{munit}
  \item AttentionGAN\cite{attentiongan}
\end{itemize}

Each model is trained using the unpaired dataset described in \ref{Unpaired-dataset}. Since the loss in the Generative Adversarial Networks training do not provide the same valuable information as in classification tasks and the difficulty of deciding whether a translation is good or bad for transferring knowledge of an object detection data set from one domain to the other, multiple checkpoints are kept from each training for later evaluation.

\begin{figure}[htp]
    \centering
    \includegraphics[width=8cm]{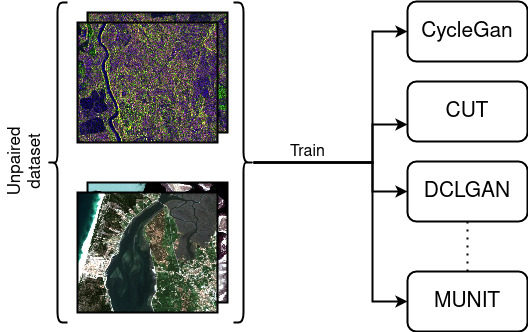}
    \caption{Image-to-image models are trained with an unpaired dataset generated with Sentinel-1 and Sentinel-2 imagery}
    \label{fig:i2i}
\end{figure}

\subsection{Validation of fake imagery}
One empirical way of measuring how much of the latent space is transferred to the destination domain is to do it indirectly by evaluating a model that extracts knowledge from the transferred dataset. In this work, this is done by training models with the translated datasets for later performing object detection in the SAR domain.

\subsubsection{Object detection datasets}
To validate the utility of the translated images, two small datasets are generated from the Sentinel-1 and Sentinel-2 imagery sources.
The resolution of both Sentinel-1 and Sentinel-2 bands used is 10m, therefore the class selected for creating the object detection datasets is \emph{ports} due to the fact that even smallest occurrences can be seen in the images with enough resolution.
For creating the object detection datasets, square patches of 870x870px with class presence are generated from both Sentinel-1 and Sentinel-2 imagery sources described in \ref{datasources}.

The sentinel-1 dataset (SAR) consists of 60 square patches of 870px side with at least one occurrence of the selected class on each patch. This dataset is used only for testing the object detection models.

The Sentinel-2 dataset (Optical) is 2.5 times bigger than the Sentinel-1 dataset (~150 samples) and is used to train the object detection models after being transformed to a fake or synthetic SAR dataset using the trained image-to-image translation models (See \ref{i2i} for details).

\begin{figure}[htp]
    \centering
    \includegraphics[width=8cm]{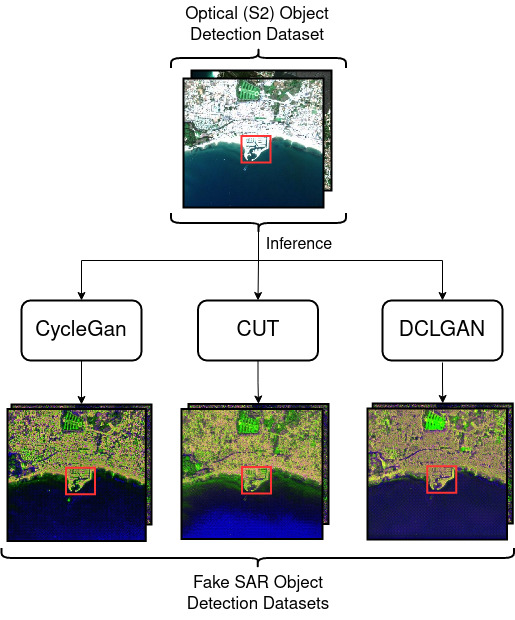}
    \caption{Fake SAR object detection datasets are created by translating the Sentinel-2 object detection dataset with the different I2I translation models}
    \label{fig:fake-sar-creation}
\end{figure}

Regarding augmentation, the situation in these datasets is completely different when compared with the datasets used in the i2i trainings. The variety and volume of the object detection ground truth is not enough to provide robustness to the detection model thus requiring an augmentation step before training the models.

Finally, the same augmentation is applied to all the datasets to perform a fair comparison later between the performances of the resulting object detection models.

\subsubsection{Model selection and training}
Two different architectures with different depths are used for the object detection task:
\begin{enumerate}
\item Faster R-CNN\cite{faster} with a Resnet50 backbone 
\item Cascade R-CNN\cite{cascade} with a Resnet101 backbone
\end{enumerate}

Both architectures have a Feature Pyramid Network (FPN) in the neck.

Transfer learning is applied in all the experiments to reduce training time and deal with such small datasets. This way, pre-trained models on COCO dataset are used as initial checkpoints in the fine-tuning process.

A proportion of 85-15\% is used for train and validation sub-datasets respectively.
Real SAR dataset is only used as a test dataset in all the object detection trainings for comparing the final performance obtained on each of them. 

Due to the difficulty in determining the best checkpoint in the I2I training for performing the Optical\textrightarrow SAR translation (See \ref{i2i}), the process of generating a fake SAR dataset from the original object detection dataset of Sentinel-2 is repeated for each I2I model and for each checkpoint saved during its training. The usability of the translated images is evaluated in all cases against the real SAR (Sentinel-1) object detection dataset.
This way, every checkpoint that produces the best results in  the object detection task are selected from each I2I translation model training.

\begin{figure*}[htp]
    \centering
    \includegraphics[width=12cm]{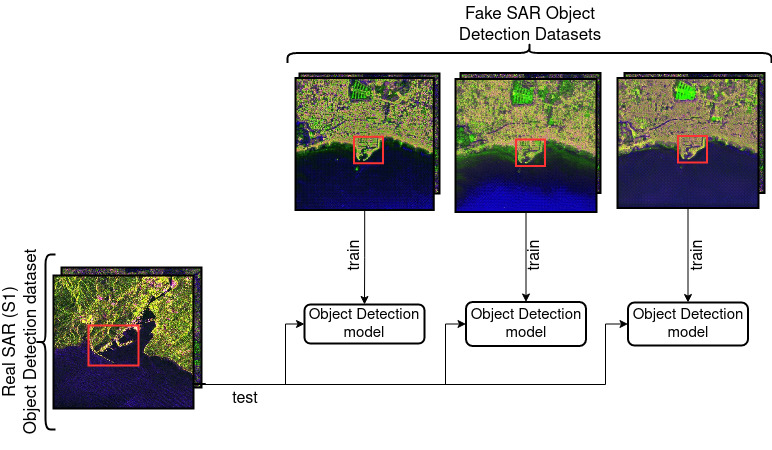}
    \caption{During training of SAR object detection models fake SAR object detection datasets (generated by the I2I translation models) are used for generating training/validation subdatasets and the models are tested against real SAR object detection dataset generated from Sentinel-1 imagery}
    \label{fig:od-train}
\end{figure*}

\subsubsection{Ensemble of models}

Stacking is a special case of ensemble methods which considers often heterogeneous weak learners that are learned in parallel and combines them by training a meta-learner to output a prediction based on the different weak learner’s predictions.

In this work, this is done using Weighted Boxes Fusion\cite{ensembles} technique. The ensemble receives  the resulting boxes from all the implied object detection models, consisting of a list of coordinates, scores and tags for each detection.

These results are weighted and later combined to generate the results.
For finding the best way of weighting the inputs of the ensemble and selecting other meta-parameters (i.e. training the meta-learner) the Optuna optimizer proposed by Akiba et al.\cite{optimizer} was used, using as performance metric a sum of average precision and recall of the resulting predictions.

\begin{figure}[htp]
    \centering
    \includegraphics[width=8cm]{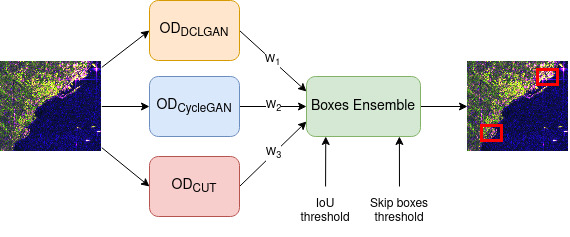}
    \caption{The boxes ensemble receives the predictions of the three object detection models and outputs a combined prediction. The ensemble adjusts the outputs using weights for the input predictions, an IoU threshold and a threshold for skipping boxes with low confidence. Weights ($W_{1}$, $W_{2}$, $W_{3}$), IoU threshold and skip boxes threshold are learned using the Optuna optimizer.}
    \label{fig:ensemble}
\end{figure}

%% file: chapters/3_results.tex
\section{Results and Discussion}

\begin{figure}[htp]
    \centering
    \includegraphics[width=8cm]{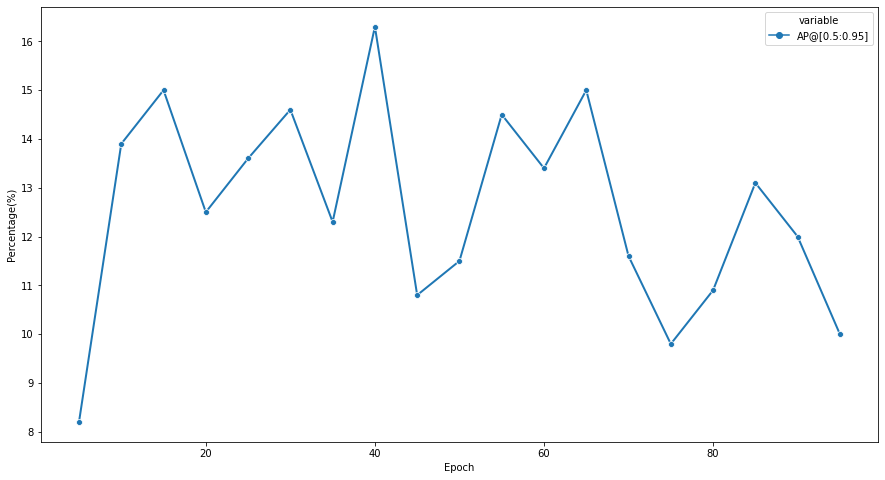}
    \caption{AP@IoU[0.5:0.95] of the different object detection models trained from datasets translated using the different checkpoints generated during the DCLGAN training}
    \label{fig:oscila}
\end{figure}

In this type of GANs, generators and discriminators are optimizing a minimax game and consequently these models do not converge and present an oscillating nature that is appreciated in all the image-to-image model trainings (example of this is shown in Fig. \ref{fig:oscila} for the DCLGAN model).
As such, it is difficult to judge whether training should stop or not. Hence, multiple checkpoints are kept for later selecting the ones that provides better results. This selection is achieved indirectly by evaluating which checkpoint produces the dataset that provides the best performance on the object detection task.

Checkpoints generated in early stages of the trainings of image-to-image models seem to work better in all implementations when used to translate the images with class presence to the SAR domain.

Probably due to the characteristics of the two domains, Council-GAN\cite{council}, MUNIT\cite{munit} and AttentionGAN\cite{attentiongan} resulting models were not capable of translating enough knowledge from one domain to the other for performing later a second level usage of the images like training an object detection model. This way, CycleGAN, CUT and DCLGAN have been selected for generating the translated object detection datasets that were later used for training the object detection models.

\begin{figure}[htp]
    \centering
    \includegraphics[width=8cm]{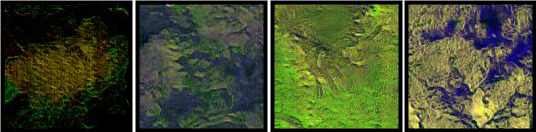}
    \caption{Examples of failed translations generated by Council-GAN, MUNIT and AttentionGAN I2I models.}
    \label{fig:failed}
\end{figure}

In Fig. \ref{fig:failed} some examples of these failed translations are shown. In all cases, although the resulting images have similarities with the destination domain, not enough details are translated from the original image to be later used as a reference in the object detection training.

\begin{table*}[htp]
  \centering
  \begin{tabular}{ |c ||c |c |c |c |c |  }
    \hline
    Object Detection Model & Object Detection Architecture & $AP_{0.5:0.95}$ & $AP_{0.5}$ & $AP_{0.75}$ & $AR_{0.5:0.95}$\\
    \hline
    \hline
    \multirow{2}{*}{OD$_{Real}$} & FasterRCNN-R50 & 3.7 & 6.9 & 2.3 & 5.4 \\
                          & CascadeRCNN-R101 & 4.5 & 9.5 & 3.5 & 5.4 \\
    \hline
    \multirow{2}{*}{OD$_{CycleGAN}$} & FasterRCNN-R50 & 16.3 & 39 & 10.9 & 20.5\\
                              & CascadeRCNN-R101 & 18.7 & 39.7 & 16.9 & 23.6\\
    \hline
    \multirow{2}{*}{OD$_{CUT}$} & FasterRCNN-R50 & 20.2 & 47.7 & 13.1 & 29.5\\
                         & CascadeRCNN-R101 & 23.2 & 57.7 & 15.3 & 29.6\\
    \hline
    \multirow{2}{*}{OD$_{DCLGAN}$} & FasterRCNN-R50 & 23.8 & 62.0 & 9.0 & 35.3\\
                            & CascadeRCNN-R101 & 20.4 & 46.3 & 11.2 & 33.9\\
    \hline
    \multirow{2}{*}{Boxes Ensemble} & FasterRCNN-R50 & 27.4 & \textbf{63.7} & \textbf{20} & \textbf{40.5}\\
                                    & CascadeRCNN-R101 & \textbf{27.5} & 63.2 & 19.5 & 38.8\\
    \hline
  \end{tabular}
  \caption{Comparison of the performance of the different object detection models evaluated against the real SAR object detection dataset generated from Sentinel-1 imagery. OD$_{Real}$ is the object detection model trained with the original Sentinel-2 dataset for object detection. OD$_{CycleGAN}$, OD$_{CUT}$, OD$_{DCLGAN}$ are the models trained with the fake SAR object detection datasets generated with CycleGAN, CUT and DCLGAN I2I models respectively. Boxes Ensemble is the ensemble model that uses outputs from OD$_{CycleGAN}$, OD$_{CUT}$, OD$_{DCLGAN}$ models.}
  \label{tab:results}
\end{table*}

The Table \ref{tab:results} shows the comparison of the performance of both types of object detection models after being trained using the different fake-SAR datasets. 

The ensemble model, that is evaluated using the outputs of the models generated from CycleGAN, CUT and DCLGAN translated datasets, shows similar performance on both architectures used of object detection, with an increase of the AP@IoU[0.5:0.95] in ~23\% when compared with the models trained with the initial S2 dataset (object detection dataset without translations between domains). 
As in other fields, the ensemble model succeeds in smartly combining the outputs from the weaker models into a stronger model. 

In this particular scenario, the different i2i techniques transfer only a part of the characteristics of the latent space to the destination domain and therefore the resulting detection models specialize in the recognition of only some of the characteristics of the objects.
By combining the outputs of the resulting object detection models, the biased knowledge of each of these models is combined, obtaining a better performance in the evaluation dataset. 
Possibly, a similar result would have been obtained as a result of applying in the first instance an i2i algorithm that performed a more reliable translation of the features of the latent space.

Despite the small size of the object detection dataset, results shows that it was perfectly possible to perform transfer learning from a model trained with optical imagery to a domain of images with a composite RGB representation of SAR products. Although the destination domain maintains the same number of channels; the textures, definition of the borders and some effects like the speckle are challenging characteristics that could have hinder the transfer learning between the two domains.

These results also confirm the hypothesis we made about the possibility of using an unsupervised image-to-image model not only to translate images from one domain to images with similar characteristics of the destination domain but also to translate at least part of the knowledge that was present in the origin domain, like the object identification as shown in this experimentation.

Finally, these results should be treated with caution as they may be invalid when using another type of representation of the SAR data or behave different with other tasks in the destination domain. In this sense, deeper experimentation should be addressed to generalize the results.

%% file: chapters/4_conclusion.tex

\section{Conclusion}

In this paper, the feasibility of using transferred data from the optical domain as a unique source of knowledge for training a model on the SAR domain is explored.

Multiple unsupervised image-to-image translation models from the state-of-the-art as well as an ensemble of the models with best results are evaluated for performing the translation between the domains.

To evaluate the usefulness of the transferred data, different object detection models are trained with this data and evaluated against a (real) SAR dataset of the same object class.

Experimental results demonstrate that a considerable number of features from the latent space are transferred with the images but in different proportions depending on the i2i algorithm. In this scenario an ensemble model is a feasible way of combining again the features translated by the different i2i algorithms.

In general, the results here demonstrate the potential value of I2I translation technique for migrating knowledge present on object detection or classification datasets in the optical domain to the SAR domain in cases where a dataset generated directly in the last is neither available nor feasible to build.

The usage of different models trained on the same origin/target domains for performing an augmentation of data from the origin domain to complement small datasets on the target domain could be a possible direction to explore as a strategy in Few-Shot learning problems.
\clearpage